\documentclass[runningheads]{llncs}
\usepackage{graphicx}
\usepackage{comment}
\usepackage{amsmath,amssymb} 
\usepackage[dvipsnames]{xcolor}
\usepackage{floatrow}
\usepackage{graphicx}
\usepackage{amssymb}
\usepackage{url,hyperref}
\usepackage{bm}
\usepackage[utf8]{inputenc} 
\usepackage[T1]{fontenc}    
\usepackage{hyperref}       
\usepackage{url}            
\usepackage{booktabs}       
\usepackage{amsfonts}       
\usepackage{nicefrac}       
\usepackage{microtype}      
\usepackage{multicol}
\usepackage{multirow}
\usepackage{graphicx}
\usepackage{gensymb}
\usepackage{booktabs,tabularx,enumitem,ragged2e}
\usepackage{subfig}
\usepackage{makecell}
\usepackage{colortbl}
\usepackage{amssymb}
\usepackage{amsmath}
\usepackage{sidecap}
\usepackage{wrapfig}
\usepackage{svg}
\usepackage[normalem]{ulem}
\useunder{\uline}{\ul}{}
\usepackage{adjustbox}
\usepackage{array}
\usepackage{floatrow}
\newfloatcommand{capbtabbox}{table}[][\FBwidth]

\usepackage{blindtext}
\newcolumntype{R}[2]{%
    >{\adjustbox{angle=#1,lap=\width-(#2)}\bgroup}%
    l%
    <{\egroup}%
}

\newcommand{\myParagraph}[1]{\textbf{#1} ---}

\begin{document}
\pagestyle{headings}
\mainmatter
\def\ECCVSubNumber{544}  

\title{Learning to plan with uncertain topological maps} 

\titlerunning{Learning to plan with uncertain topological maps}
%
\author{Edward Beeching\inst{1}\and
Jilles Dibangoye\inst{1}\and
Olivier Simonin\inst{1}
\and
Christian Wolf\inst{2}
}

\authorrunning{E. Beeching et al.}
%
\institute{INRIA Chroma team, CITI Lab. INSA Lyon, France.
\url{https://team.inria.fr/chroma/en/} \and
Université de Lyon, INSA-Lyon, LIRIS, CNRS, France.\\
\email{\{firstname.lastname\}@insa-lyon.fr}}
\maketitle
\textbf{Project page}\href{edbeeching.github.io/papers/learning_to_plan}{ https://edbeeching.github.io/papers/learning\_to\_plan}

\begin{abstract}
\noindent
We train an agent to navigate in 3D environments using a hierarchical strategy including a high-level graph based planner and a local policy. Our main contribution is a data driven learning based approach for planning under uncertainty in topological maps, requiring an estimate of shortest paths in valued graphs with a probabilistic structure. Whereas classical symbolic algorithms achieve optimal results on noise-less topologies, or optimal results in a probabilistic sense on graphs with probabilistic structure, we aim to show that machine learning can overcome missing information in the graph by taking into account rich high-dimensional node features, for instance visual information available at each location of the map. Compared to purely learned neural white box algorithms, we structure our neural model with an inductive bias for dynamic programming based shortest path algorithms, and we show that a particular parameterization of our neural model corresponds to the Bellman-Ford algorithm.
By performing an empirical analysis of our method in simulated photo-realistic 3D environments,  we demonstrate that the inclusion of visual features in the learned neural planner outperforms classical symbolic solutions for graph based planning. 
\end{abstract}

\section{Introduction}
\noindent
A critical part of intelligence is navigation, memory and planning. An animal that is able to store and recall pertinent information about their environment is likely to exceed the performance of an animal whose behavior is purely reactive. 
Many control and navigation problems in partially observed 3D environments involve long term dependencies and planning. It has been shown that humans and other animals navigate through the use of waypoints combined with a local locomotion policy \cite{Wang2002-WANHSR,gupta2017unifying}. 
In this work, we mimic this strategy by proposing a hierarchical planner, which performs high-level long term planning using an uncertain topological map (a valued graph including visual features) combined with a local RL-based policy navigating between high-level waypoints proposed by the graph planner.
Our main contribution is a way to combine symbolic planning with machine learning, and we look to structure a neural network architecture to incorporate landmark based planning in unseen 3D environments.

\begin{figure}[t]
    \centering
    \includegraphics[width=1.0\textwidth]{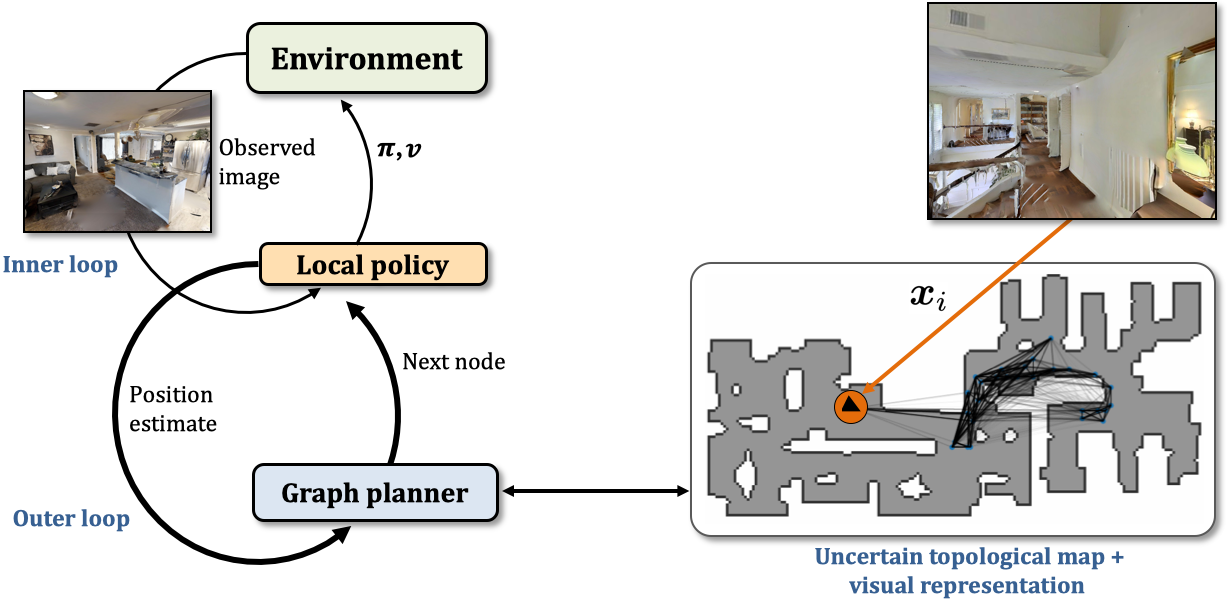}
    \caption{\label{fig:teaser}A trained agent navigates to a goal location with a hierarchical planner. A high-level  planner proposes new target nodes in a topological map (a graph), which are used as an objective for a local point-goal policy. The graph is estimated from an explorative rollout and, as such, uncertain: the opacities of the edges correspond to estimations of connectivity between nodes (darker lines = higher confidence). In this example we observe a low probability of connection between the node at the agent's position (orange) and its nearest neighbor, whereas from the visual observation associated to the node we can see there is a traversable space between the two nodes.} 
\end{figure}

When solving visual navigation tasks, biological or artificial agents require an internal representation of the environment if they want to solve more complex tasks than random exploration. We target a scenario where an agent is trained on a large-scale set of 3D environments to learn to reason on planning and navigation. When faced with a previously unseen environment, the agent is given the opportunity to build a representation by doing an explorative rollout from a previously learned explorative policy. It can then exploit this internal representation in subsequent visual navigation tasks. This corresponds to many realistic situations, where robots are deployed to indoor environments and are allowed to familiarize themselves before performing their tasks \cite{savinov2018semiparametric}.

Our agent constructs an imperfect topological map of its environment in the form of a graph, where nodes correspond to places and valued edges to connections. Edges are assigned two different values, the first one being spatial distances, the second one being probabilities indicating whether it is possible to navigate between the two nodes. Nodes are also assigned rich visual features extracted from images taken at the corresponding places in the environment. After deployment, the agent faces visual navigation tasks requiring it to find a specific location in the environment provided by a set of images corresponding to different viewpoints, extending the task proposed in \cite{zhu2017target}.  The objective is to identify the goal location in the internal representation, and to provide an estimate for the shortest path to it. The main difficulty we address here is the fact that this path is an estimate only, since the ground truth path is not available during testing / deployment. 

Whilst planning in graphs with known connectivity has been solved for many decades \cite{dijkstra1959note,bellman1958routing}, planning under uncertainty remains an ongoing area of research. Whereas optimal results in a probabilistic sense 
exist for graphs with probabilistic connectivity, we aim to show that machine learning can overcome missing information in the graph by taking into account rich high-dimensional node features, in particular features extracted from image observations associated with specific nodes. We train a graph neural network in a fully supervised way to predict estimates of the shortest path, using vision to overcome uncertainty in the connectivity information. We present a new variant of graph neural networks imbued with specific inductive bias, and we show that this structure can be parameterized to fallback to the classical Bellman-Ford algorithm.

Figure \ref{fig:teaser} illustrates the hierarchical planner: a neural graph based planner runs an outer loop providing estimates for next way-point on a graph, which are used as target nodes for a local RL-based policy running an inner loop and providing feedback to high-level planner on reached locations. Both planners take into account visual features, either stored in the graph (graph based planner), or directly as observations provided by the environment (local policy). The two planners are trained separately --- the graph based planner in a fully supervised way from ground truth graphs, the local policy with RL and a point-goal strategy.

\noindent
This work makes the following contributions:
\begin{itemize}
    \item A hierarchical model combining high-level graph based planning with a local point goal policy for robot navigation;
    \item A trainable high-level neural planner which combines an uncertain topological map (graph) with rich node features to learn to estimate shortest paths in noisy and unknown environments. 
    \item A variant of graph networks encoding inductive bias inspired by dynamic programming-based shortest path algorithms.
    \item We evaluate the performance of the method in challenging and visually realistic 3D environments and show that it outperforms optimal symbolic planning on noisy topological maps.
\end{itemize}
\begin{figure}[t]
    \centering
    \includegraphics[width=\textwidth]{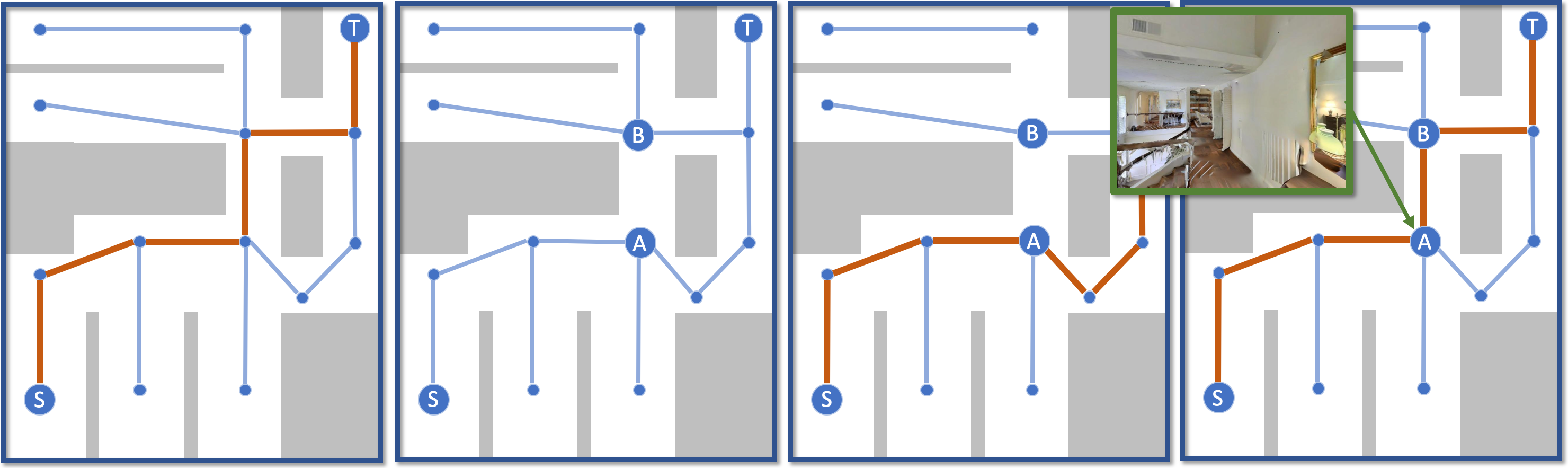} \\
    (a) \hspace{2.5cm} (b) \hspace{2.5cm} (c) \hspace{2.5cm} (d) 
    \caption{\label{fig:4graphs}Illustration of the different types of solutions to the high-level graph planning problem: (a) the ground truth graph (unavailable during testing) with the shortest path from node $S$ to node $T$ in red; (b) the uncertain graph available during test time. This graph is fully connected and for each edge a connection probability is available. For clarity we here show only edges where the connection probability is above a threshold. The edge from $A{\rightarrow}B$ is wrongly estimated as not connected; (c) an ``optimal'' path taking into account both probabilities and distances; (d) A learned shortest path, where the visual features at node $A$ indicate passage to node $B$. We supervise a network to predict the GT path (a).}
\end{figure}

\section{Related work}
\noindent
\textbf{Classical planning and graph search} ---
A large body of work is available on classical planning on graphs, notable references include \cite{lavalle2006planning,Remolina2004TowardsAG}. In robotics, there have been a number of works applying classical planning in topological maps for indoor robot navigation, for instance \cite{shatkay1997learning,thrun1998learning}.

\noindent
\textbf{Planning under imperfect information ---} 
In many realistic robotic problems, the current state of the world is unknown. 
Though sensor observations provide measurements about the current state of the world, these measurements are usually incomplete or noisy because of disturbances that distort their values.
Planning problems that face these issues are referred to as planning problems under imperfect information.
Research on this topic has a long history, which can be traced back to the seminal work by \cite{aastrom1965optimal} presenting the first non-trivial exact dynamic programming algorithm for partially observable Markov decision processes (POMDPs). While there are other models \cite[chap 12]{lavalle2006planning}, POMDPs emerged as the standard framework to formalize and solve (single-agent) sequential decision-making problems with imperfect information about the state of the world \cite{kaelbling1998planning}. Since the agent does not have access to the actual state of the world, it acts based solely on its entire history of actions and observations, or the corresponding belief state, \emph{i.e., } the posterior probability distribution over the states given the history \cite{aastrom1965optimal,smallwood1973optimal}. Approaches for finding optimal solutions have been intensively investigated in the 2000s, ranging from dynamic programming \cite{kaelbling1998planning} to heuristic search methods \cite{smith2004heuristic,kurniawati2008sarsop}. Key to these approaches is the idea that one can recast the original problem into a continuous-state fully observable Markov decision process, where states are belief states or histories \cite{aastrom1965optimal}. Doing so allows theory and algorithm that applies for MDPs to also apply to POMDPs, albeit in much larger (and possibly continuous) state space. Another significant result of this literature is proof that the optimal value function is a piece-wise linear and convex function of the belief states, which allows the design of algorithms with faster rates of convergence \cite{smallwood1973optimal}.  For a thorough discussion on existing solvers for POMDPs, the reader can refer to \cite{shani2013survey}. 

\noindent
\textbf{Deep Reinforcement Learning} ---
The field of Deep Reinforcement Learning (RL) has gained attention with successes on board games  \cite{Silver1140} and Atari games \cite{mnih2015humanlevel}. Recent works have applied Deep RL for the control of an agent in 3D environments \cite{Mirowski2016LearningEnvironments} \cite{Jaderberg2016ReinforcementTasks}, exploring the use of auxiliary tasks such as depth prediction, loop detection and reward prediction to accelerate learning. Other recent work uses street-view scenes to train an agent to navigate in city environments \cite{MirowskiLearningMap}. To infer long term dependencies and store pertinent information about the partially observable environment, network architectures typically incorporate recurrent memory such as Gated Recurrent Units \cite{ChungGatedNetworks} or Long Short-Term Memory \cite{Hochreiter1997LONGMEMORY}.
Extensions to memory based neural approaches began with Neural Turing Machines \cite{graves2014neural} and Differentiable Neural Computers \cite{graves2016hybrid},
and have since been adapted to expand the capacity of Deep RL agents \cite{Wayne2018UnsupervisedPM}.
Spatially structured  memory architectures have been shown to augment an agent's performance in 3D environments and are broadly split into two categories: metric maps which discretize the environment into a grid based structure and topological maps which produce node embeddings at key points in the environment. Research in learning to use a metric map is extensive and includes spatially structured memory \cite{parisotto2018}, Neural SLAM based approaches \cite{DBLP:journals/corr/ZhangTBBL17} and approaches  incorporating projective geometry and neural memory \cite{cogmap2017,Bhatti2016PlayingDW}, these techniques are combined, extended and evaluated in \cite{beeching2020egomap}. Other notable works in include that of Value Iteration Networks (VIN) \cite{tamar2016value} which approximate the value iteration algorithm with a CNN, applied planning in small fully observable state spaces (grid worlds). While VIN and our work structure planners, VINs use convolutions to approximate classical value iteration, while we use a graph representation and a novel GNN architecture with recurrent updates to approximate the Bellman-Ford algorithm. \cite{karkus2017qmdpnet} plans under uncertainty in partially observable gridworld environments. Here uncertainty refers to POMPs, the classical QMDP algorithm is used as inductive bias for a neural network, whereas in our work uncertainty is over node connectivity in a graph constructed in a previously unseen environment. \cite{srinivas2018universal} which is applied in observable state spaces to learn a forward model in a latent space to plan appropriate actions; they are not hierarchical, are not graph-based and do not appear to plan under uncertainty. Similar to ours, they are applied to goal driven problems.

Research combining learning, navigation in 3D environments and topological representations has been limited in recent years with notable works being \cite{savinov2018semiparametric} who create graph a through random exploration in ViZDoom RL environment \cite{Kempka2017ViZDoom:Learning}. \cite{NIPS2019_9660} also performs planning in 3D environments on a graph-based structure created from randomly sampled observations, with node distances estimated with value estimates. The downside of these approaches is that in order to generalize to an unseen environment, many random samples must be taken in order to populate the graph.

\noindent
\textbf{Graph neural networks} ---
Graph Neural Networks (GNN) are deep networks that operate on graphs directly. They
have recently shown great promise in domains such as knowledge graphs \cite{schlichtkrull2018modeling}, chemical analysis \cite{gilmer2017neural}, protein interactions \cite{fout2017protein}, physics simulations \cite{battaglia2016interaction} and social network analysis \cite{kipf2016semi}. These types of architectures enable learning from both node features and graph connectivity. Several review papers have covered graph neural networks in great detail \cite{bronstein2017geometric,battaglia2018relational,wu2019comprehensive,zhou2018graph}. GNNs have been applied to shortest path planning in travelling salesmen problems \cite{li2018combinatorial,joshi2019efficient} and it has been reasoned that they can approximate optimal symbolic planning algorithms such as the Bellman-Ford algorithm \cite{xu2019can}. This work applies a novel variant of GNN in order to solve approximate planning problems, where classical methods may struggle to deal with uncertainty.

\section{Hierarchical navigation with uncertain graphs}
\noindent
We train an agent to navigate in a 3D visual environment and to exploit an internal representation, which it is allowed to obtain from an explorative rollout before the episode. Our objective is image goal, i.e. target-driven navigation to a location which is provided through a (visual) image. We extend the task introduced in \cite{zhu2017target} by generalizing to unseen environment configurations without the need to retrain the agent for a novel environment.

From the explorative rollout obtained with an agent trained with RL, which is further described in section \ref{sec:graphcreation}, we create an uncertain topological map covering the environment, i.e. a valued graph $\mathcal{G}{=}\{\mathcal{V},\bm{V},\bm{E},\bm{L},\bm{D}\}$, where $\mathcal{V}{=}\{1,\dots N\}$ is a set of nodes, $\bm{V}$ is a $K{\times}N$ matrix of rich visual node features of dimensions $K$, $\bm{E}\in[0,1]^{N{\times}N}$ is a set of edge probabilities where $\bm{E}_{i,j}$ is the probability of having an edge between nodes $i$ and $j$, $\bm{L}$ is a matrix of node locations and $\bm{D}$ is a distance matrix, where $\bm{D}_{i,j}$ is a distance between nodes $i$ and $j$. While $\bm{D}$ encodes a distance in a path planning sense, $\bm{E}$ encodes the probability of $j$ being directly accessible from $i$ with obstructions. The uncertainty encoded by this probability can be considered to be a combination of aleatory variability, i.e. uncertainty associated with natural randomness of the environment, as well as epistemic uncertainty, i.e. uncertainty associated with variability in computational models for estimating the graph, in our case the explorative policy trained with RL and taking into account visual observations.

Once the topological map is obtained, the objective of the agent at each episode is to navigate to a location given an image, which is provided as additional observation at each time step. The agent acts in 3D environments like Habitat\cite{habitat19iccv} (see section \ref{sec:experiments}), receiving images of the environment as observations and predicting actions from a discrete space (\emph{forward}, \emph{turn left 10~$^{\circ}$}, \emph{turn right 10~$^{\circ}$}). We propose a hierarchical planner performing actions at two different levels:
\begin{description}
\item[A high-level graph based planner] that operates on longer time scale $\tau$ and iteratively proposes new point-goals nodes $p_g^\tau$ that are predicted, by a Graph Neural Network, to be on the shortest path from the agent to the estimated location of the target image.
\item[A local policy] that has been trained to navigate to a local point-goal $p_g^\tau$, which has been provided by the high-level policy. The local policy operates for a maximum of $m$ time-steps, where $m$ is a hyper-parameter, set to 10. The agent has been trained with an additional \textit{STOP} action, so that it can learn to terminate the local policy in the case that it reaches $p_g^\tau$ in under $m$ steps.
\end{description}
The two planners communicate through estimated locations, the graph planner indicating the next waypoint to the local policy as a location, and the local policy (after termination) providing an estimate of its reached location back to the high-level planner. The planner updates its current node estimate as the nearest neighboring node and planning continues.

\subsection{High-level planning with uncertain graphs}
\noindent
The objective of the high-level planner is to estimate the shortest path from the current position  $S{\in}\mathcal{V}$ in the graph to a terminal node $T{\in}\mathcal{V}$, whose identity is estimated as the node whose visual features are closest to the target image in cosine distance. Planning takes into account the distances between nodes encoded in $\bm{D}$ as well as estimated edge connectivity encoded in $\bm{E}$. As an edge $(i,j)$ may have a large connection probability $\bm{E}_{i,j}$ but still be obstructed in reality, the goal is to learn a trainable planner parameterized by parameters $\theta$, which takes into account visual features $\bm{V}$ to overcome the uncertainty in the graph connectivity. To this end, we assume the ground truth connectivity $\bm{E}^*$ available during training only. Figure \ref{fig:4graphs} illustrates the different types of solutions this problem admits: the optimal shortest path is only available on ground truth data (Figure \ref{fig:4graphs}a), the objective is to use the noisy uncertain graph (Figure \ref{fig:4graphs}b) and provide an estimate of the optimal solution taking into account visual features (Figure \ref{fig:4graphs}d). This is unlike the optimal solution in a probabilistic sense calculated from a symbolic algorithm (Figure \ref{fig:4graphs}c).

We propose a trainable planner, which consists of a novel graph neural network architecture with dedicated inductive bias for planning. 
Akin to graph networks \cite{Battaglia_2018_arXiv}, the node embeddings are updated with messages over the edges, which propagate information over the full graph.
While it has been shown that graph networks can be trained to perform planning \cite{Xu2019NNReason}, we aim to closely mimic the structure of the Bellman-Ford algorithm and we embue the planner with additional inductive bias and a supervised objective to explicitly learn to calculate shortest paths from data. To this end, each node $i$ of the graph is assigned an embedding $\bm{x_i}=[ \bm{v_i}, \bm{e_i}, t_i, \bm{d_i}, \bm{s_i}]$ where $\bm{v_i}$ are visual features from the memory matrix $\bm{V}$, $t_i$ is a boolean value indicating if the node is the target, $\bm{e_i}$ are the edge connection probabilities from node $i$ to all other nodes, $\bm{d_i}$ are the distances for node $i$ to all other nodes, $\bm{s_i}$ is a one hot vector identifying the node (part of the identity matrix $\bm{I}$). 

We motivate our proposed neural model with the following objective: the planner should be able to exploit information contained in the graph connectivity, but also in the visual features, to be able to find the shortest path from a given current node to a given target node. As with classical planning algorithms, it will thus eventually be required to keep for each node a latent representation of the bound $d_i$ on the shortest distance as well as information on the identity of the outgoing edge to the neighbor lying on the shortest path, the predecessor function $\Pi(i)$. Known algorithms (Dijstra, Bellman-Ford) perform iterative updates of these variables $(d_i,\Pi_i)$ by comparing them with neighboring nodes and the corresponding inter-node distances, updating the bound $d_i$ and $\Pi_i$ when a shorter path is found than the current one. This is usually done by iterating over the successors of a given node $i$.

\begin{figure}[t]
    \centering
    \includegraphics[width=0.62\textwidth]{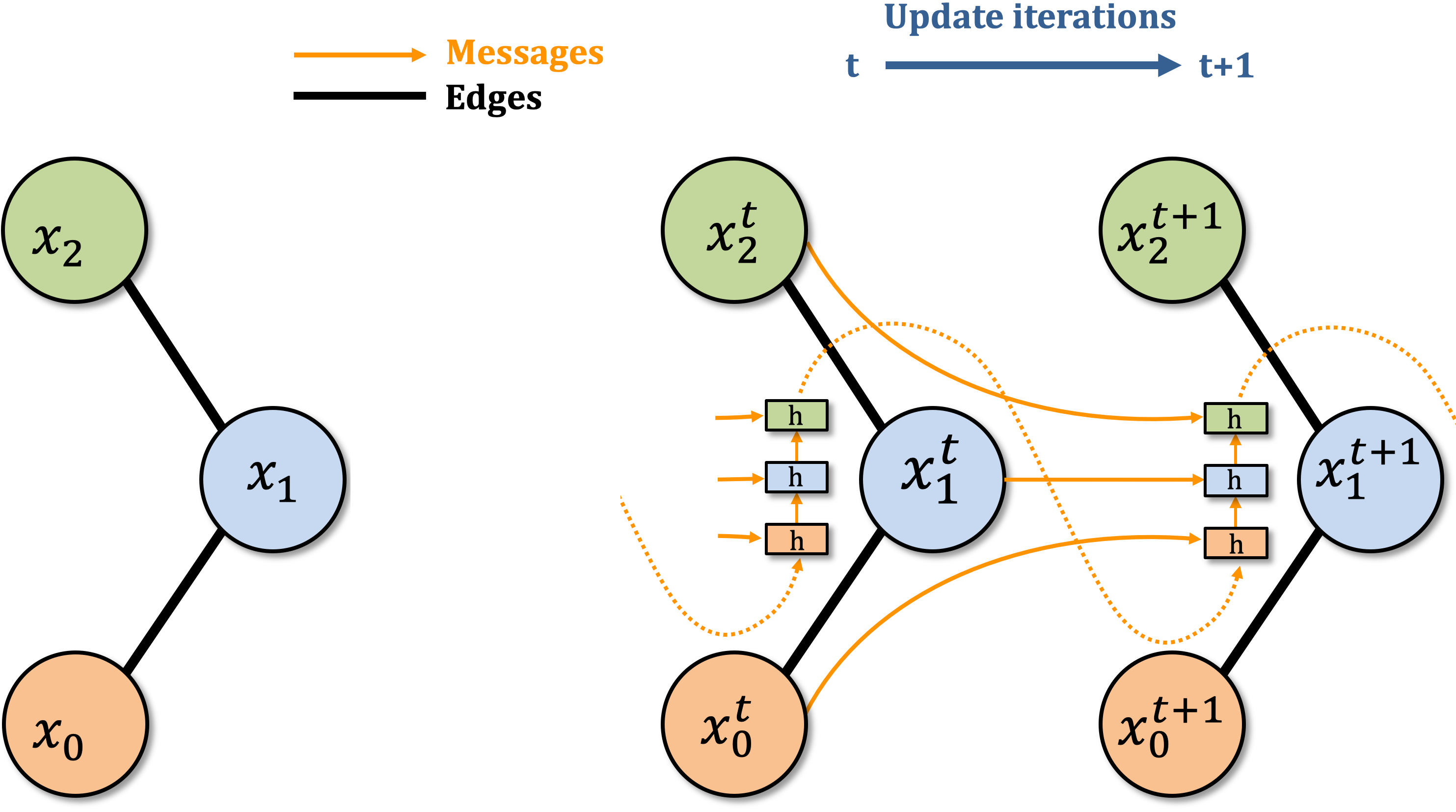} \\
    (a) \hspace{5.52cm} (b) \hspace{2cm}
    \caption{(a) An example graph; (b) One iteration of the neural graph planner's message passing and bound update. Incoming messages from neighbors are serialized and fed through a recurrent unit, which creates inductive bias for learning minima necessary for bound updates.}
    \label{fig:oneiteration}
\end{figure}

In our trained model, these variables are not made explicit, but they are supposed to be learned as a unique vectorial latent representation for each node $i$ in the form of an internal state $\bm{r_i}$, which generally holds current information on the reasoning of the agent. The input to each iteration of the graph network is, for each node $i$, the node embedding $\bm{x_i}$, and the node state $\bm{r}_i$, which we concatenate to form a single node vector $\bm{n}_i$:
$$
\bm{n_i} = [ \bm{x}_i, \bm{r}_i ] = [ \bm{v_i}, \bm{e_i}, t_i, \bm{d_i}, \bm{s_i}, \bm{r_i}]
$$
As classically done in graph neural networks, this representation is updated iteratively by exchanging messages between nodes in the form of trainable functions. The messages and trainable functions of our model are given as follows, illustrated in Figure \ref{fig:oneiteration}, and will be motivated in detail further below.
\begin{align} 
\bm{m}_{i,j} & = 
      \bm{W}_1  [ \bm{n}_i,\bm{n}_j ] \odot 
\sigma(\bm{W_2} [ \bm{n}_i,\bm{n}_j ] ) \label{eq:gatedlinear} \\
\bm{r'}_i &= \phi^{r\leftarrow h}(\{\bm{m}_{i,j}\}_{\forall j}, \bm{h}_i) \label{eq:gru} 
\end{align}
Here, $\odot$ is the Hadamard product, $\bm{W}_.$ are weight matrices, and $\bm{r'}_i$ is the updated latent representation after one round of  updates. The features $\bm{x}_i$ do not change during these operations. 

Equation (\ref{eq:gatedlinear}) is inspired from gated linear layers \cite{dauphin2017language}, and enables each node to identify whether it is the target, and update its representation of the bound. We use gated linear layers in order to provide the network with the capacity to update bound estimates for its neighbors.

Equation (\ref{eq:gru}) integrates messages from all neighbors $j$ of node $i$, updating its latent representation. Since planning requires this step to update internal bounds on shortest paths, akin to shortest path algorithms that rely on dynamic programming, we serialize the updates from different neighbors into a sequence of updates, which allows the network to learn to calculate minimum functions on bound estimates. In particular, we model this through a recurrent network in a Gated Recurrent Unit variant \cite{chung2014empirical}, using a hidden state vector $\bm{h}_i$ associated to each node $i$. The step is structured to mimic the min operation of the Bellman-Ford algorithm (see section \ref{sec:relationtobf} for details on this equivalence).

Equation (\ref{eq:gru}) can thus be rewritten in more detail as follows: Going sequentially over the different neighbors $j$ of node $i$, the hidden state $\bm{h}_i$ is updated as follows:
\begin{equation}
\bm{h}_i^{[j]} = \bm{W}_3 \bm{m}_{i,j} + \bm{W}_4 \bm{h}_i^{[j-1]}
\label{eq:grudetail}
\end{equation}
For simplicity, we omitted the gating equations of GRUs and presented a single layer GRU. In practice we include all gating operations and use a stacked GRU with two layers. The output of the recurrent unit is a non-linear function of the last hidden state, providing the new latent value $\bm{r}'_i$:
\begin{equation}
\bm{r}'_i = MLP(\bm{h}_i^N)
\end{equation}
where $MLP$ is a two-layer neural network with ReLU activations.

The above messages are exchanged and accumulated for $k$ steps where $k$ is a hyper-parameter which should be at least the largest span of the graphs in the dataset. 
The action distribution $f_A(\bm{r_i})$ is then estimated for each node in the graph as a linear mapping of the node embeddings followed by a softmax activation function. 
\begin{equation}
\bm{A}_i = f_A(\bm{r_i}) = \textrm{softmax}(\bm{W} \bm{r_i})
\label{eq:action}
\end{equation}

\subsection{Relations to optimal symbolic planners}
\label{sec:relationtobf}
\noindent
As mentioned before, our neural planner could in theory be instantiated with a specific set of network parameters such that it corresponds to a known symbolic planner calculating an optimal path in a certain sense. To illustrate the relationship of the network structure, in particular the recurrent nature of the graph updates, we will layout details for the case where the planner performs the estimation of a shortest path given the distance matrix and ignoring the uncertainty information --- an adaptation to an optimal planner in the probabilistic sense can be done in a straightforward manner. To avoid misunderstandings, we insist that the reasoning developed in this sub section is for illustration and general understanding of the chosen inductive network bias only, the real network parameters are fully trained with supervised learning as explained in section \ref{sec:training}.

Handcrafting a parameterization requires imposing a structure on the node state $\bm{r}_i$, which otherwise is a learned representation. In our case, the node state will be composed of the bound $b_i$ on the shortest path from the given node to the target node (a scalar), and the current estimate $\Pi_i$ of the identity of predecessor node of node $i$ w.r.t. the shortest path, which can be represented as a 1-in-K encoded vector indicating a distribution over nodes. 

Standard Bellman-Ford symbolic bound updates iteratively update the bound for a given node $i$ by examining all its neighbors $j$ and checking whether a shorter path can be found passing through neighbor $j$.
This can be written in a sequential form s.t. the bound gets updated iterating through the neighbors $j{=}1{\dots}J_i$ of node $i$:
\begin{equation}
\begin{array}{ll}
b_i^{[0]} &=  b_i \\
b_i^{[j]} &=  \min (b_i^{[j-1]}, b_j + d_{ij}) \\
b_i' & =  b_i^{[J_i]}
\end{array}
\label{eq:bfupdate}
\end{equation}
where $b_i$ is the bound before the round of updates for node $i$, and $b'_i$ is the bound after the round of updates for node $i$.

In our neural formulation, the message updates given in equation (\ref{eq:gru}), further developed in (\ref{eq:grudetail}), mimic the Bellman-Ford bound update given in Equations (\ref{eq:bfupdate}). This provided motivation for our choice of a recurrent neural network in the graph neural network, as we require the update of the recurrent state $h_i^j$ in Equation (\ref{eq:grudetail}) to be able to perform a minimum operation and an arg min operation (or differentiable approximations of min and arg min).

\subsection{Graph creation from explorative rollouts}\label{sec:graphcreation}
\noindent
Graphs were generated during the initial rollout from an exploratory policy trained with Reinforcement Learning. During training, the agent interacts with training environments and receives RGB-D image observations calculated as a projection from the 3D environment. The agent is trained to explore the environment and to maximize coverage, i.e. to visit as much space as possible as quickly as possible similar to \cite{Chaplot2020Learning,chen2018learning}.

To learn to estimate the graph connectivity, we add an auxiliary loss to the agent's objective function, $f_{link}(\bm{o_i}, \bm{o_j}, \bm{h_i})$ which is trained to classify whether two locations are in line of sight of each other, conditioned on the visual features $\bm{o_i}, \bm{o_j} $ from the two locations and the agent's hidden state $\bm{h_i}$. Node features were calculated with a CNN \cite{Lecun1998Gradient-BasedMethod}. Ground truth line of sight measurements were computed by 2D ray tracing on an occupancy map of each environment. 
In order to limit the size of the graph to a maximum number of nodes $k$, we aim to maximize each node's coverage of the environment using a Gaussian kernel function. At each time step a new node is observed by the agent, previous node positions are compared with a Gaussian kernel function (eq. \ref{eq:kernel}) in order to identify the index of the most redundant node $r$, which is removed from the graph and replaced with the new node, node connectivities are then recomputed with $f_{link}(.)$, where $\bm{L}_i$ is the location of node $i$.
\begin{equation}\label{eq:kernel}
r = \arg\ \min_i 
\left ( \sum_j K(\bm{L}_i,\bm{L}_j) \right )
, \quad \quad 
K(\bm{v}, \bm{v'})= \textrm{exp}(\frac{- \| \bm{v}-\bm{v'} \|^2}{2\sigma^2}),
\end{equation}
\section{Training}\label{sec:training}
\noindent
\textbf{The high-level graph based planner ---} is trained in a purely supervised way. We generate ground truth labels by running a symbolic algorithm (Dijkstra \cite{dijkstra1959note}) on a set of valued ground truth training graphs described with the method detailed in Section \ref{sec:graphcreation}. In particular, the supervised training algorithm takes as input \emph{uncertain/noisy} graphs, which include visual features, and is supervised to learn to produce paths, which are calculated from known ground truth graphs unavailable during test time. During training we treat path planning as a classification problem where for a given target, each node must learn to predict the subsequent node on the optimal path to the target. 

Formally for each node $i$ we predict a distribution $\bm{A_i}$ and aim to match a ground-truth distribution $\bm{A_i^*}$, which is a one-hot vector, minimizing cross entropy loss $
\mathcal{L}(\bm{A},\bm{A^*})=-\sum_{i=1}^{n} {\bm{A}^*_{i} \log \bm{A}_{i}}$.

We augment training with a novel version of mod-drop \cite{neverova2015moddrop}, a training algorithm for multi-modal data, which drops modalities probabilistically during training. In our case, during training we extend the node connection probabilities in the input with the ground truth node adjacencies and mask either the probabilities or the adjacencies with a probability of 50\%, during training we linearly taper the masking probability from 50\% to 100\% over the first 250 epochs. This ensures that the final model requires only connection probabilities, but the reasoning performed during message passing and recurrent updates can be bootstrapped from the ground truth adjacency matrix. Training curves on unseen validation data are shown in figure \ref{fig:moddrop}.

\noindent
\textbf{The local policy ---} is a recurrent version of AtariNet \cite{Mnih2015Human-levelLearning} with two output heads for the action distribution and value estimates. The network was trained with a reinforcement learning algorithm Proximal Policy Optimization (PPO) \cite{Schulman2017ProximalAlgorithms} to navigate with discrete actions to a local point-goal. Point-goals were generated to be within 5m of the spawn location of agent. A dense reward was provided that corresponds to a decrease in geodesic distance to the target, a large reward (10.0) was provided when the agent reached the target and the \textit{STOP} action was used. The episode was terminated when either the \textit{STOP} action was used or after 500 time-steps. A small negative reward of -0.01 was given at each time-step to encourage the agent to complete the task quickly.

\begin{figure} [t] \centering
    \subfloat[\label{fig:dijkstra_thresholded} ]{\includegraphics[width=0.49\textwidth]{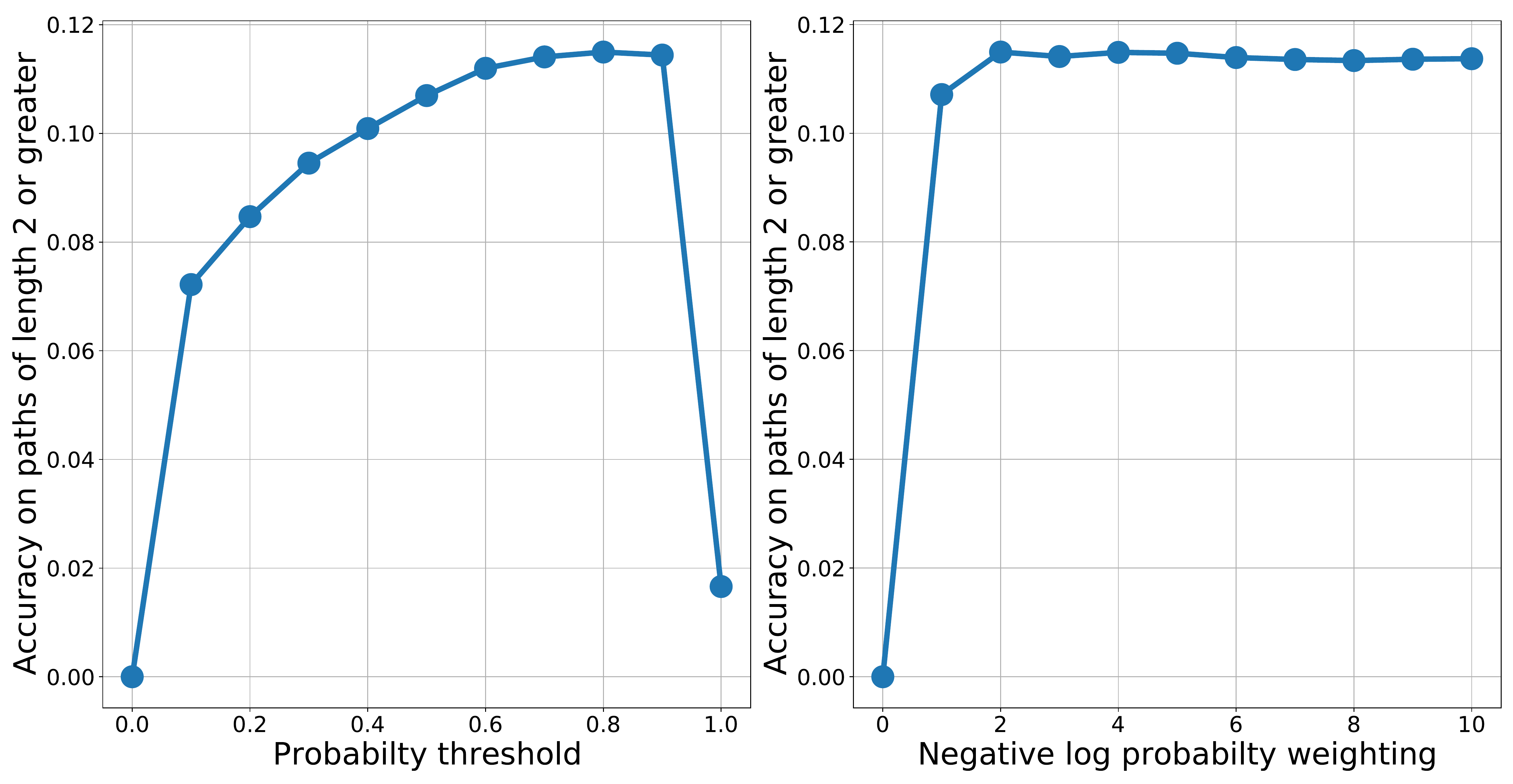}}
    \hspace{0.1cm}
    \subfloat[\label{fig:flink_training}]{\includegraphics[width=0.49\textwidth]{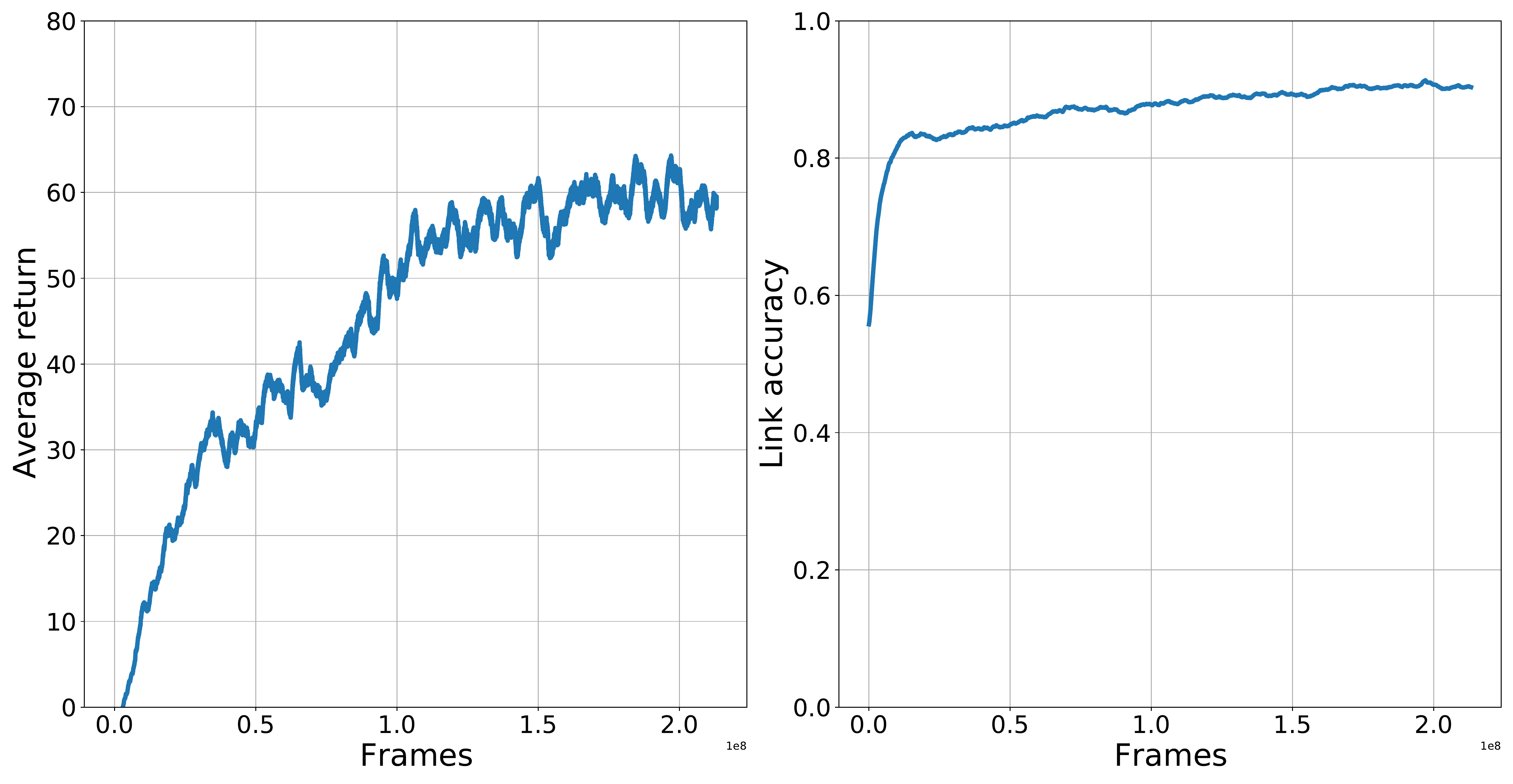}}
    \caption{(a) Symbolic baselines Dijkstra on - left: thresholded probs., right: cost function(\ref{eq:dijk_custom}). (b) Left: Average return. Right: Accuracy of line of sight predictions.}
\end{figure}

\begin{figure}[t]%
 \centering
 \subfloat[]{\includegraphics[width=0.47\textwidth]{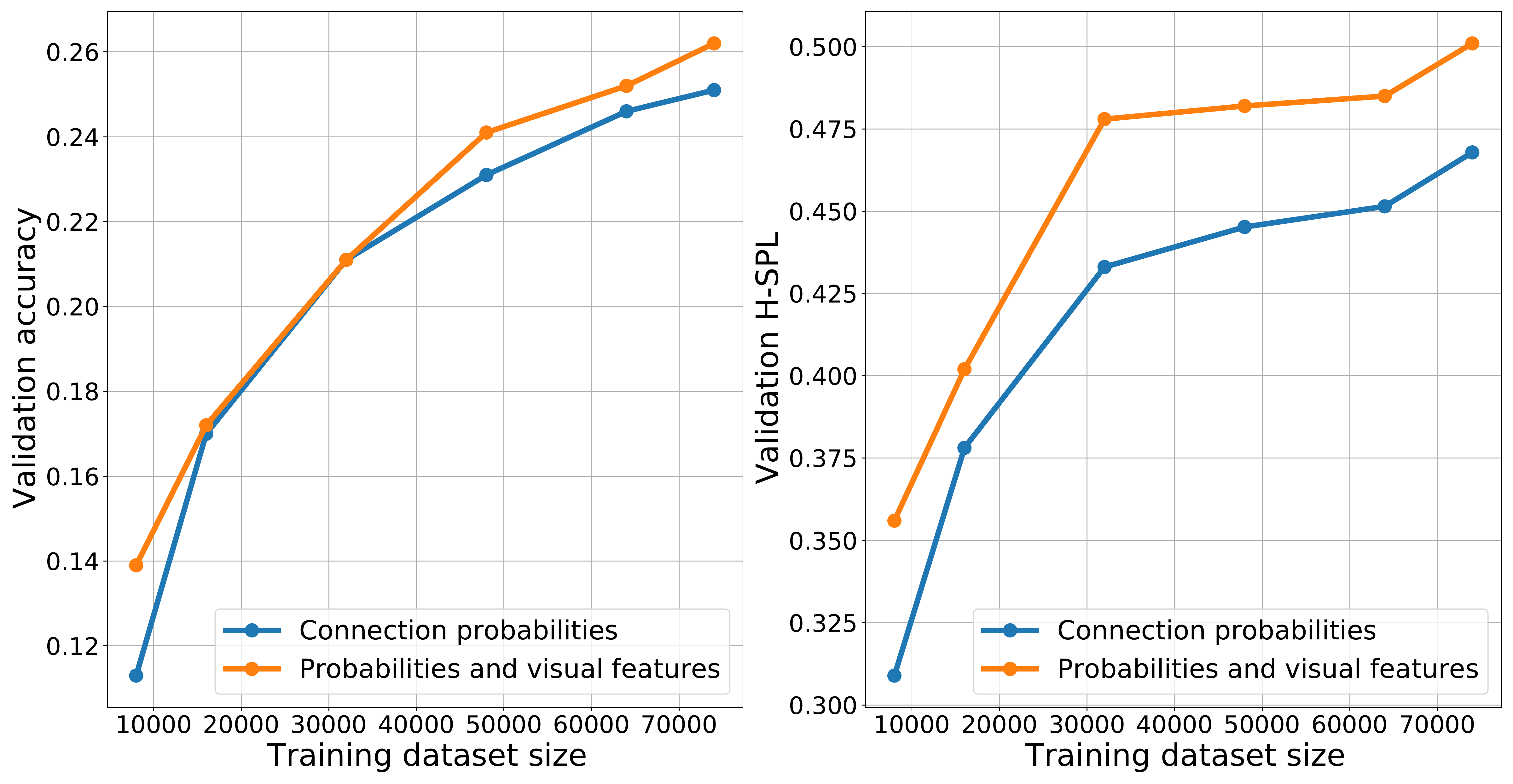}\label{fig:dataset_size}}%
 \subfloat[]{\includegraphics[bb = 0 0 1400 800,clip=true,width=0.48\textwidth]{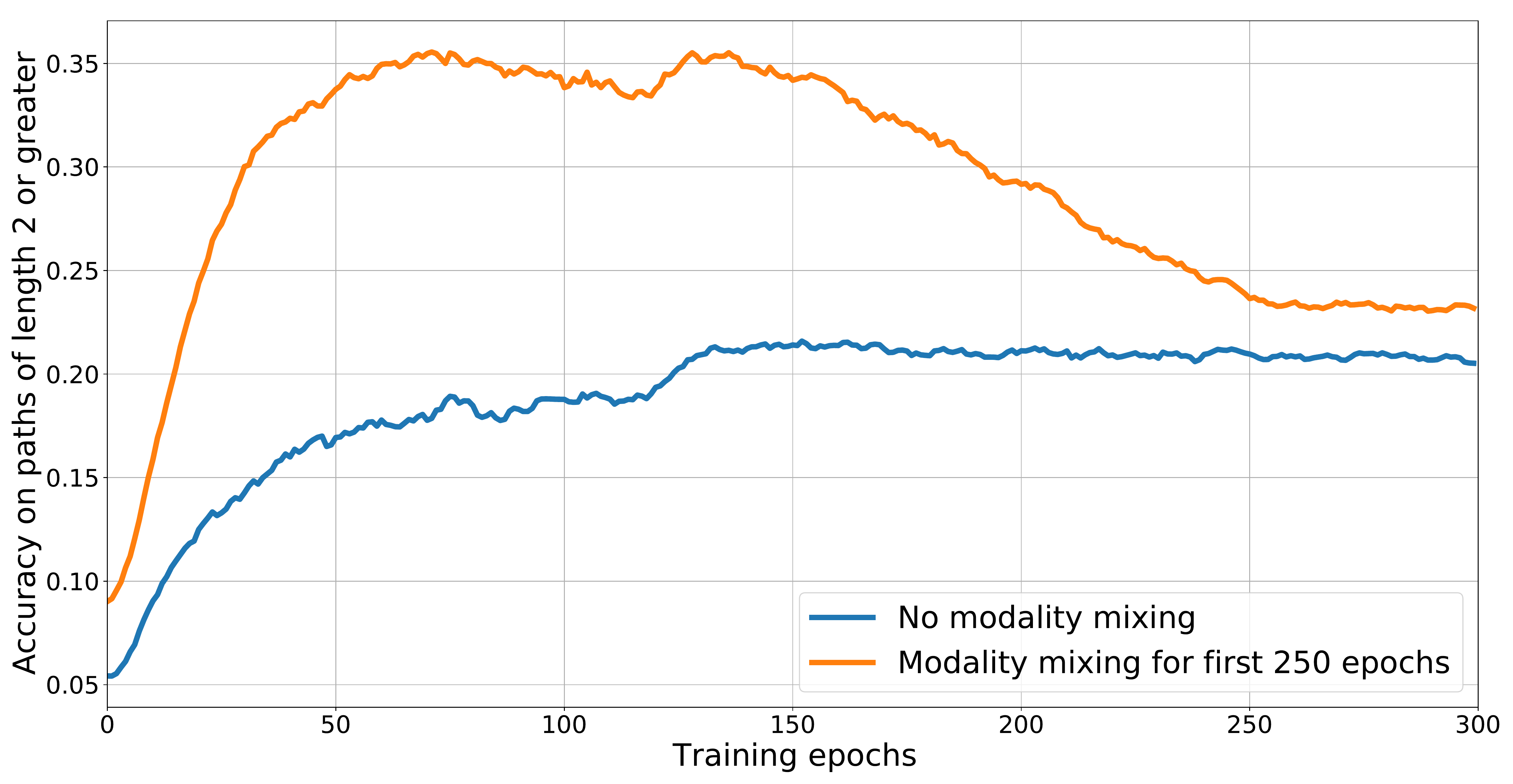}\label{fig:moddrop}}\\
 \caption{(a) Accuracy and H-SPL with increasing size of data when the GNN is trained with and without visual features. (b) Modality mixing (GT connections and probabilities), we observe a 2.3\% improvement over single modality training.}%
\end{figure}

\noindent
\textbf{The explorative policy for graph creation ---} is trained with PPO \cite{schulman2017proximal}. We aim to maximize coverage that is within the field of view of the agent. We create an occupancy grid of the environment with a grid spacing of 10cm. The first time a cell is observed the agent receives a reward of 0.1. A cell is considered to observable if it is free space, within 3m of the agent and in the field of view of the agent. Agent performance is shown in Figure \ref{fig:flink_training}.

\begin{table}[t] \centering
\caption{Reporting H-SPL and accuracy of the neural planner's predictions in unseen environments; neural planner trained with 72,000 graphs}
\subfloat[\label{tab:results_probs}Results on uncertain graphs]
{
\resizebox{0.45\textwidth}{!}{%
\begin{tabular}{l|cc}
\hline
\textbf{Method} & \textbf{Acc} & \textbf{H-SPL} \\ \hline
Symbolic (threshold) & 0.114 & 0.184 \\
Symbolic (custom cost) & 0.115 & 0.269 \\
Neural  (w/o visual) & 0.251 & 0.468 \\
Neural  (w visual) & \textbf{0.262} & \textbf{0.501} \\ \hline
\end{tabular}%
}}
\subfloat[\label{tab:results_gt}Results on ground truth graphs]
{
\resizebox{0.45\textwidth}{!}{%
\begin{tabular}{l|cc}
\hline
\textbf{Method} & \textbf{Acc } & \textbf{H-SPL } \\ 
\hline
Symbolic (GT) & \textbf{1.00} &\textbf{1.00} \\
Neural planner (GT) & 0.921 & 0.983 \\ \hline
\end{tabular}%
}}
\end{table}

\begin{figure}[t]
    \centering
    \includegraphics[width=0.9\textwidth]{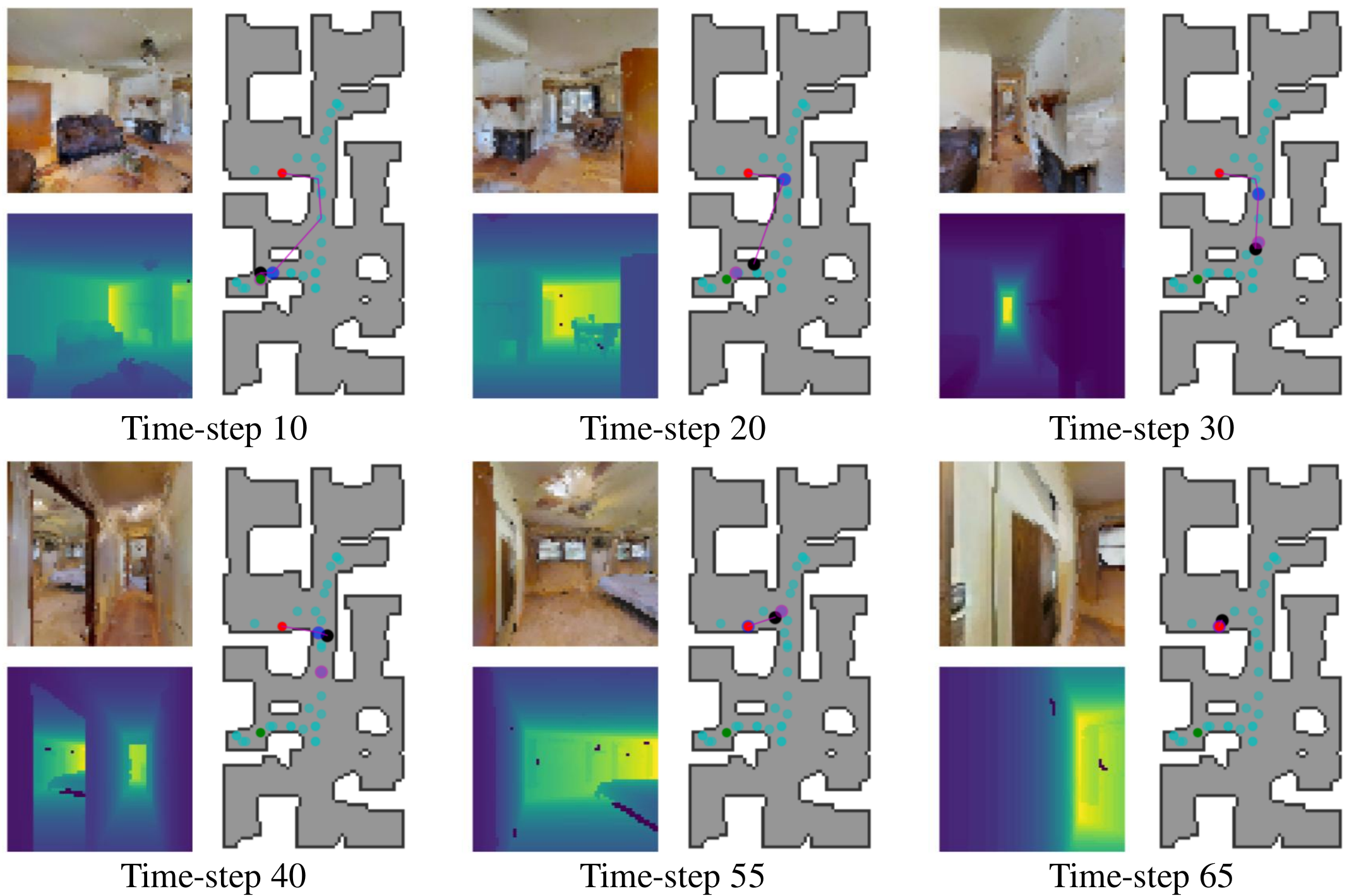}
    \caption{
    Six time-steps from a rollout of the hierarchical planner (graph+local) in an unseen testing environment. For each time-step: left -- RGB-D observation, right -- map of the environment (unseen) with  \textcolor{Cyan}{\underline{graph nodes}},  \textcolor{OliveGreen}{\underline{source node}},   \textcolor{red}{\underline{target node}},  \textcolor{black}{\underline{agent position}}(black),  \textcolor{Fuchsia}{\underline{nearest neighbour}} to the agent, \textcolor{Blue}{\underline{local point-goal}} provided by the high level planner and  \textcolor{Fuchsia}{\underline{planned path}}. Further examples can be found in the supplementary material, including failure cases.
    }
    \label{fig:planning_example}
\end{figure}

\section{Experiments}
\label{sec:experiments}
We evaluated our method in simulated 3D environments, in particular the Habitat  \cite{habitat19iccv} simulator with the visually realistic Gibson dataset \cite{xiazamirhe2018gibsonenv}. During training, the agent interacts with 72 different training environments from Gibson, where each environment corresponds to a different apartment or house, and receives as input observation an observed RGB-D image. 
We evaluate our method on a set of 16 held out environments that were unseen during training by either the local policy, the exploratory policy or the high-level neural planner. 
\subsection{High-level graph-based planner}
\noindent
The neural planner was implemented in PyTorch \cite{NEURIPS2019_9015}, the hyper-parameters are given in the supplementary material. 
We compare two metrics, accuracy of prediction of the next way-point along the optimal path and the SPL metric \cite{anderson2018evaluation}, both for paths of length two or greater. As we evaluate SPL for both the high level planner and the hierarchical planner-controller, we refer to the high-level planner's SPL as H-SPL to avoid ambiguity.

\noindent
\textbf{Symbolic baselines ---}
We compare the neural planner to two symbolic baselines, both of which reason on the uncertain graph only, without taking into account rich node features. While these baselines are \emph{``optimal''} with respect to their respective objective functions, they are optimal with respect to the amount of information available to them, which is uncertain:
 \emph{(i) Thresholding} ---
In order to generate non-probabilistic edge connections, we threshold the connection probabilities with values ranging from 0-1 in steps of 0.1. After threshholding the graph, path planning was performed with Dijkstra's algorithm;
\emph{(ii) A custom cost function} for Dijkstra's algorithm weighting distances and probabilities:
\begin{equation}\label{eq:dijk_custom}
cost(i,j) = \bm{D}_{i,j} - \lambda \log(\bm{E}_{i,j})
\end{equation}
We vary the weighting $\lambda$ in order to control the trade-off of distance and connection probability. In the limit where $\lambda$ is 0, the graph is a fully connected graph, whereas high values of $\lambda$ would lead to finding the most probable path.

Results of both symbolic baselines for varying hyper-parameters are shown in Figure \ref{fig:dijkstra_thresholded}, we observe that they perform poorly under uncertainty. In both cases we aim to evaluate the accuracy of their predictions with respect to the symbolic baseline on the ground truth graph, i.e. Dijkstra on the shortest path. As graphs can contain many source-target pairs that are within 1 step, we report accuracy on source-target pairs separated by at least 2 steps.

\textbf{Image driven recurrent baseline ---} We also compare to an end-to-end RL approach where the current observation and target image are provided to a CNN based RL agent trained from reward. The agent architecture is a siamese CNN with a recurrent GRU. We train with a dense reward of improvement in geodesic distance between the agent and the target, and provide a reward of 10 when the agent reaches the goal. We used PPO, and trained for 200 M environment frames.

In Table \ref{tab:results_probs}, we compare the neural planner,the two symbolic baselines and the recurrent baseline. We can see, that even without visual features, the neural planner is able to outperform the ``optimal'' symbolic baselines. This can be explained with the fact, that the baselines optimize a fixed criterion, whereas the neural planner can learn to exploit patterns in the connection probability matrix $\bm{E}$ to infer valuable information on shortest ground truth path. The gap further increases when the neural planner can use visual features. The positive impact of modality mixing (see section \ref{sec:training}) is shown in Figure \ref{fig:moddrop}.

As a sanity check, table \ref{tab:results_gt} compares the optimal symbolic planner against the neural planner trained with ground truth adjacencies provided as input. We observe that the results of the neural planner are close to optimum in this case. 

We evaluated our approach with different amounts of training data, ranging from 8,000 graphs to 74,000 training graphs (Figure \ref{fig:dataset_size}). Note that one training graph spawns 32$\times$32 possible source-target combinations, leading to a maximum amount of 75,000,000 training instances.

\begin{table}[t]
\centering
\caption{Performance of the hierarchical graph planner \& local policy}
\label{tab:hier_acc_spl}
{ \small
\begin{tabular}{l|cc}
\hline
\textbf{Method: Planner + Local policy} & \textbf{Success rate} & \textbf{SPL} \\ 
\hline \hline
\emph{Graph oracle (optimal point-goals, not comparable)} & 0.963 &	0.882
 \\ 
\hline
Random & 0.152 & 0.111 \\
Recurrent Image-goal agent & 0.548 & 0.248 \\
Symbolic (threshold) & 0.621 & 0.527 \\
Symbolic (custom cost) & 0.707 & 0.585 \\
Neural planner (sampling) & 0.966 & 0.796 \\
Neural planner (deterministic) & \textbf{0.983} & \textbf{0.877} \\ \hline
\end{tabular}%
}
\end{table}
\subsection{Hierarchical planning and control (topological \& local policy)}
We evaluated the neural graph planner coupled with the local policy. For a given episode, the graph planner estimates the next node in the path to a target image and provides its location to the local policy, which executes for $m$ time-steps. The planner then re-plans from the nearest neighbor to the agent's current position, this back and forth process of planning and navigating continues until either the agent reaches the target or 500 low-level time-steps have been conducted. We report accuracy as percentage of runs completed successfully and SPL in table \ref{tab:hier_acc_spl}, albeit measured on low-level trajectories as opposed to graph space. We combine the local policy with various graph planners, and can see that the neural graph planners greatly outperform the symbolic baselines.
We perform two evaluations of the neural planner; a deterministic evaluation where point-goals are chosen with the argmax of the $\bm{A}$ distribution and a non-deterministic one by sampling from $\bm{A}$. The motivation is that by sampling, the planner can escape from local minima and loops created by errors in approximation. This is confirmed when studying rollouts from the agents, and also quantitatively through the performances shown in table \ref{tab:hier_acc_spl}.  A visualization of steps from an episode is shown in Figure \ref{fig:planning_example}, where in step 10 we can see that navigation is robust w.r.t. local errors in planning (the purple line crossing white non-traversable space).
\subsection{Ablation: Effect of chosen inductive bias}
\noindent
As developed in sections 3.1 and 3.2, our graph based planner includes a particular inductive bias, which allows it to represent the Bellman-Ford algorithm for the calculation of shortest or best paths. This bias is implemented as a recurrent model (a GRU) running sequentially over the message passing procedures, as illustrated in Figure \ref{fig:oneiteration}.

\begin{figure}[t]
    \centering
    \includegraphics[width=0.8\textwidth]{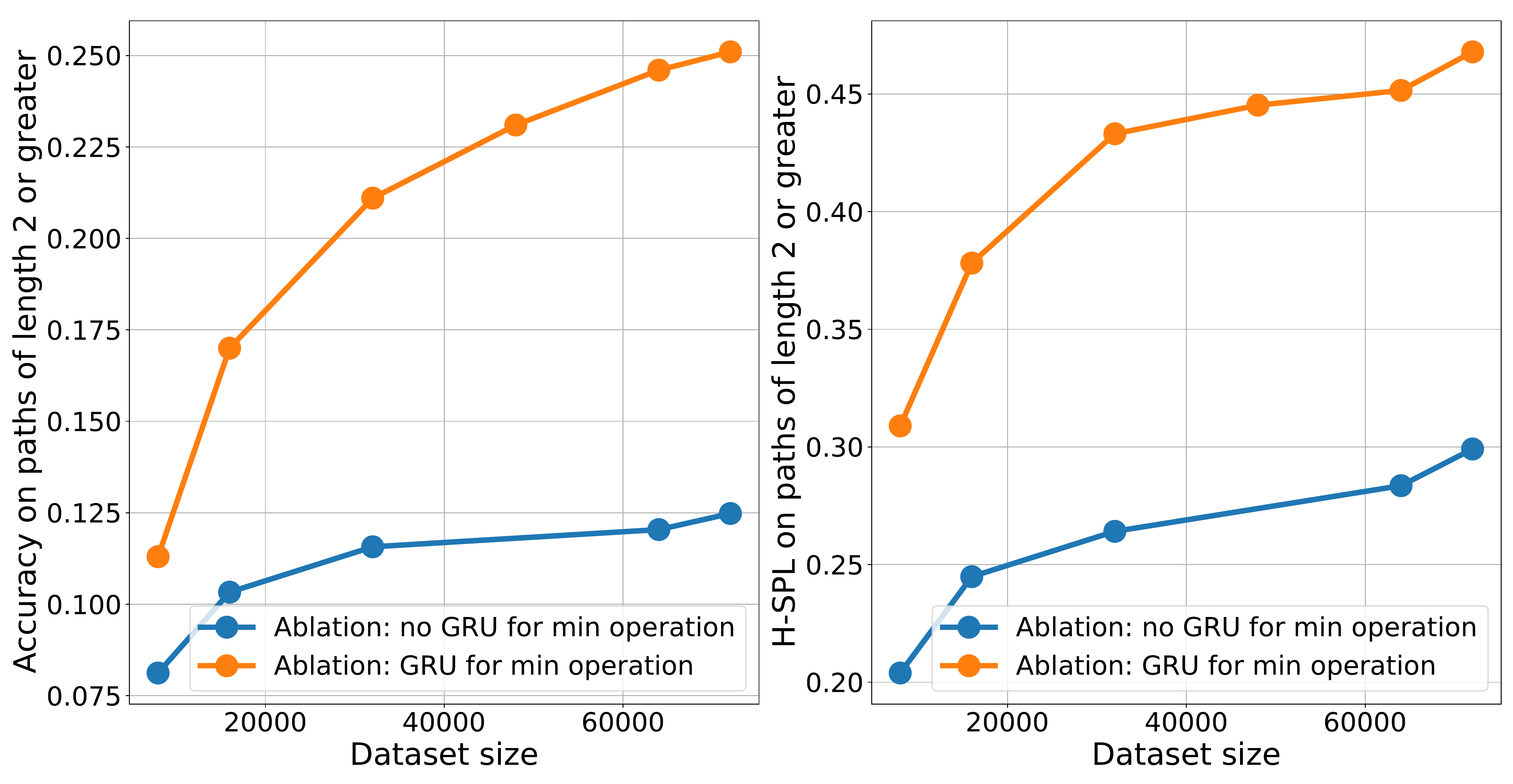}
    \caption{
    Ablation of the addition of a GRU for the accumulation of incoming messages. This recurrent unit was added to ensure that the model could represent the Bellman-Ford algorithm.
    }
    \label{fig:gru_ablation}
\end{figure}

Figure \ref{fig:gru_ablation} ablates the effect of this additional bias as a function of data sizes ranging from 8,000 to 74,000 example graphs, each one evaluated with 32$^2$=1,024 different combinations of starting and end points.

The differences are substantial, and we can see that our model is able to exploit increasing amounts of data and translates them into gains in performance, whereas standard graph convolutional networks don’t --- we conjecture that they lack in structure allowing them to pick up the required reasoning.

\section{Conclusion}
\noindent
We demonstrated that path planning algorithms can be approximated with learning when structured in a manner that is akin to classical path planning algorithms. We have performed an empirical analysis of the proposed solution in photo-realistic 3D environments and have shown that in uncertain environments graph neural networks can outperform their symbolic counterparts by incorporating rich visual features as part of their planning procedure. Our method can be used to augment a vision based agent with the ability to form long term plans under uncertainty in novel environments, without a priori knowledge of the particular environment. We have analysed the empirical performance of the neural planning algorithm with a variety of dataset sizes, shown that the high-level planner can be coupled with a low-level policy and evaluated the hierarchical performance on an image-goal task. 

\myParagraph{Acknowledgements}
 This work was funded by grant Deepvision (ANR-15-CE23-0029, STPGP479356-15), a joint French/Canadian call by ANR \& NSERC; Compute was provided by the CNRS/IN2P3 Computing Center (Lyon, France), and by GENCI-IDRIS (Grant 2019-100964).

%
%
\bibliographystyle{splncs04}
\bibliography{refs}


\newpage
\section{Supplementary material}
\subsection{Example graphs}

\begin{figure*}[h]
    \centering
    \includegraphics[width=1\textwidth]{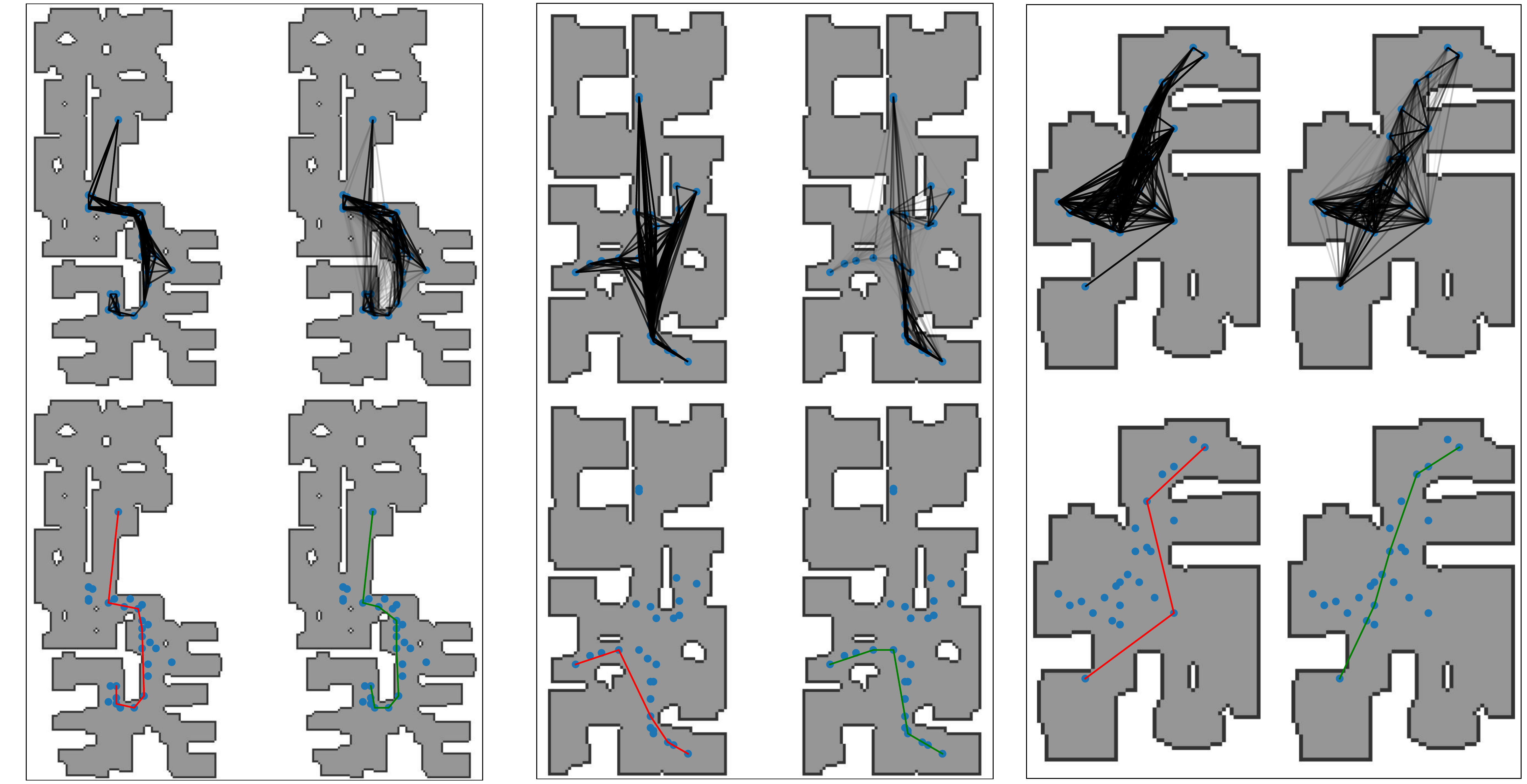}
    \caption{
    Examples of top down maps and graphs from three environments. For each environment - Top-left: topological map with ground truth connectivities, top-right: topological map with connection probabilities estimated by the learned $f_{link}$ function with line opacity corresponding to the link probability, bottom-left: ground truth shortest path between a source an target node, bottom-right: shortest path estimated by the Graph Neural Planner. Note the very bottom-right prediction connects two nodes that are not connected in the ground truth, this example would not be counted as a valid path during evaluation of the SPL metric.
    }
    \label{fig:graph_examples}
\end{figure*}

\noindent
Figure \ref{fig:graph_examples} shows example graphs extracted from three different environments extracted with the method described in section 3.3 of the main paper.

\begin{figure}[p]
    \centering
    \includegraphics[width=0.9\textwidth]{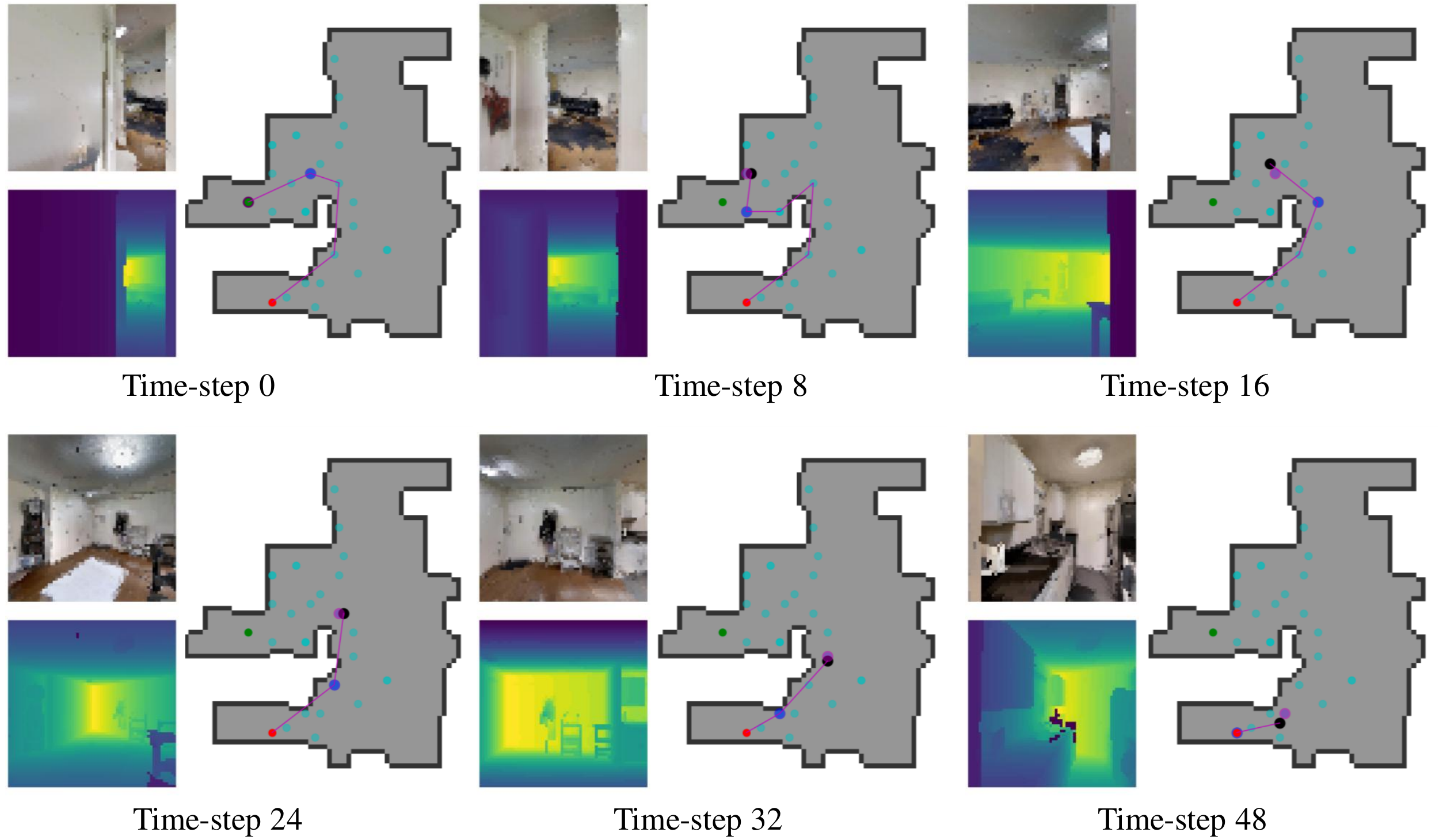}
    \caption{
    Six time-steps from a rollout of the hierarchical planner (graph+local) in an unseen testing environment. For each time-step: left -- RGB-D observation, right -- map of the environment (unseen) with  \textcolor{Cyan}{\underline{graph nodes}},  \textcolor{OliveGreen}{\underline{source node}},   \textcolor{red}{\underline{target node}},  \textcolor{black}{\underline{agent position}}(black),  \textcolor{Fuchsia}{\underline{nearest neighbour}} to the agent, \textcolor{Blue}{\underline{local point-goal}} provided by the high level planner and  \textcolor{Fuchsia}{\underline{planned path}}. 
    }
    \label{fig:planning_example2}
\end{figure}
\begin{figure}[p]
    \centering
    \includegraphics[width=0.9\textwidth]{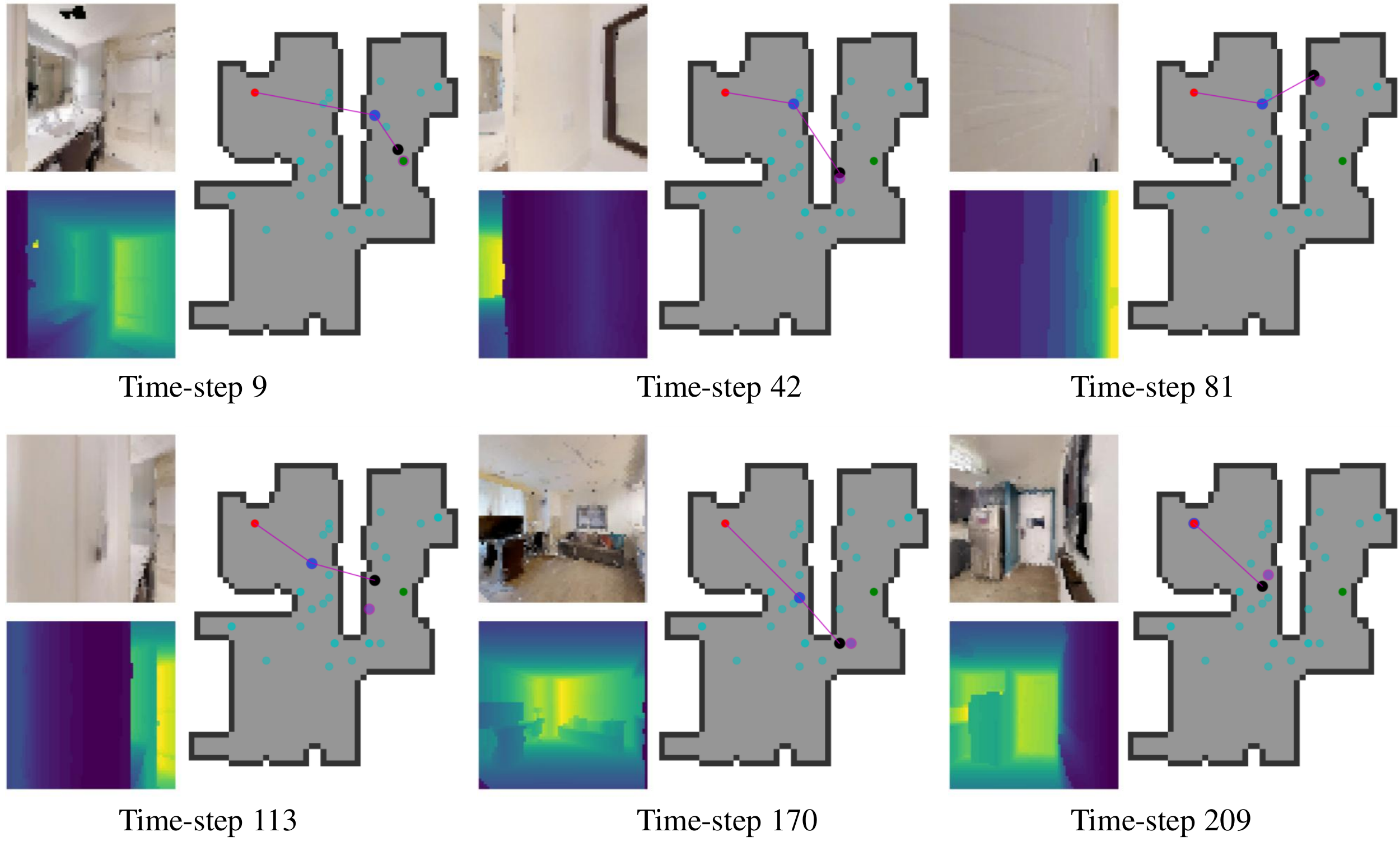}
    \caption{
    Failure case - Six time-steps from a rollout of the hierarchical planner (graph+local) in an unseen testing environment. For each time-step: left -- RGB-D observation, right -- map of the environment (unseen) with  \textcolor{Cyan}{\underline{graph nodes}},  \textcolor{OliveGreen}{\underline{source node}},   \textcolor{red}{\underline{target node}},  \textcolor{black}{\underline{agent position}}(black),  \textcolor{Fuchsia}{\underline{nearest neighbour}} to the agent, \textcolor{Blue}{\underline{local point-goal}} provided by the high level planner and  \textcolor{Fuchsia}{\underline{planned path}}.
    }
    \label{fig:planning_example3}
\end{figure}

\subsection{Examples of Graph planner trajectories}

\noindent
Figures \ref{fig:planning_example2} and \ref{fig:planning_example3} of this document show additional rollouts of the hierarchical planner, complementary to figure 6 of the main paper.

\subsection{Effect of the length $m$ of the local policy }
\begin{figure}[t]
    \centering
    \includegraphics[width=1.0\textwidth]{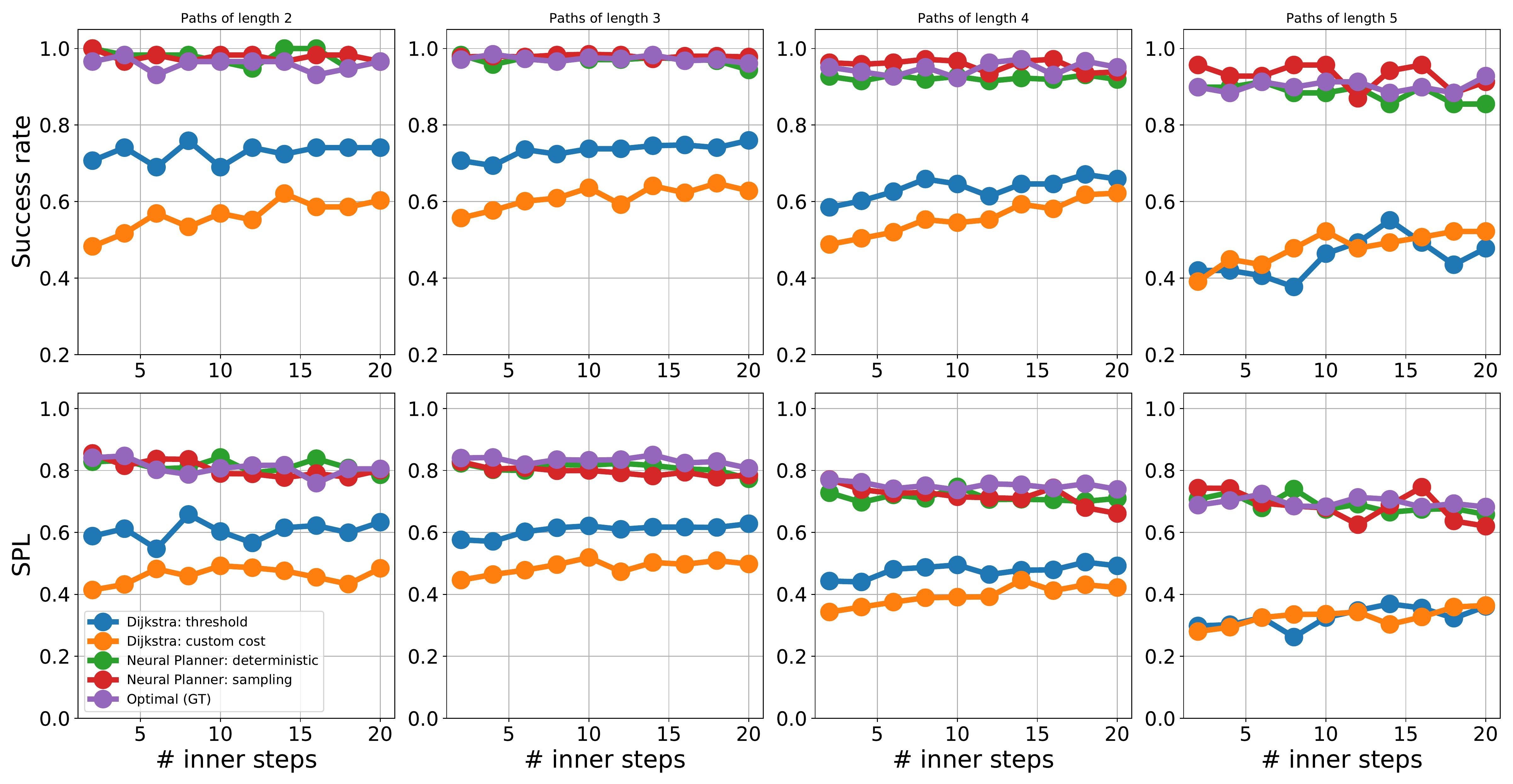}
    \caption{
    Performance of the hierarchical agent with varied low level inner loop steps for a range of high-level path lengths.
    }
    \label{fig:inner_steps}n
\end{figure}

\noindent
In figure \ref{fig:inner_steps} of this document we show the effect of the parameter $m$ of the local policy, described in section 3, page 6, of the main document, when evaluated on a limited random subset of the validation data (1,200 problem instances). The $m$ parameter limits the maximum number of steps the local policy can take before giving control back to the high-level graph planner. We recall that the local policy can also decide to terminate the inner loop earlier through an explicit \emph{STOP} action. We see that performance of the planner and policy is comparable up to 20 time-steps, which means the computationally costly planning step can performed less frequently than the low level control of the local policy, without a reduction in performance.
\subsection{Hyper-parameters}

\noindent
Table
\ref{tab:hyperparameters} provides the hyper-parameters for the three different neural models used in this work: 
\begin{itemize}
    \item the explorative policy used to create the graphs, trained through RL;
    \item  the graph based high-level planner, trained in a supervised way, and
    \item the local policy, trained through RL.
\end{itemize} 

\begin{table}[t] 
\centering
\caption{Hyper-parameters for the exploratory policy, node linkage function and Neural Planner}
\label{tab:hyperparameters}
{\small
\begin{tabular}{ll}
\hline
\multicolumn{2}{c}{\textbf{Exploratory policy and $F_{link}$}} \\ \hline
Simulator resolution & 64$\times$64\\
optimizer & Adam: betas=(0.9, 0.999), eps=1e-5 \\
learning rate & 2.50e-04 \\
weight decay & 0.0 \\
parallel agents & 16 \\
GRU hidden size & 512 \\
Entropy coef & 0.001 \\
Advantage normalization & True \\
Generalized Advantage Estimation & True \\
Minibatch size & 4 (trajectories)\\
PPO CLIP & 0.1\\
num environment steps & 200 M \\
TBPTT & 128 \\ \hline
\multicolumn{2}{c}{\textbf{Neural Planner}} \\ \hline
optimizer & Adam: betas=(0.9, 0.999), eps=1e-8 \\
learning rate & 0.001 \\
weight decay & 0.0001 \\
batch size & 32 \\
num epochs & 500 \\
dataset size & 36,000 \\
GRU size & 256 \\
Feature size & 512 \\
Learning rate decay: 0.1 every 120 epochs \\
GNN steps & 6 \\
Multilayer GRU Depth & 2 \\
Gradient norm clipping & 2.0 \\\hline
\multicolumn{2}{c}{\textbf{Local policy}} \\ \hline
Simulator resolution & 64$\times$64\\
optimizer & Adam: betas=(0.9, 0.999), eps=1e-8 \\
learning rate & 2.50e-04 \\
weight decay & 0.0 \\
parallel agents & 16 \\
GRU hidden size & 512 \\
Entropy coef & 0.001 \\
Advantage normalization & True \\
Generalized Advantage Estimation & True \\
Minibatch size & 4 (trajectories)\\
PPO CLIP & 0.1\\
num environment steps & 200 M \\
TBPTT & 128 \\ \hline

\end{tabular}%
}
\end{table}

\end{document}